\documentclass[11pt,a4paper]{article}
\usepackage[hyperref]{emnlp2020}
\usepackage{times}
\usepackage{latexsym}
\usepackage{graphics}
\usepackage{amsmath}
\usepackage{tikz-cd}
\usepackage{caption}
\usepackage{float}

\newcommand{\argmin}[1]{\underset{#1}{\operatorname{arg}\,\operatorname{min}}\;}
\newcommand{\argmax}[1]{\underset{#1}{\operatorname{arg}\,\operatorname{max}}\;}
\usepackage{microtype}

\usepackage{environ}
\usepackage{csquotes}
\usepackage{amsmath}
\usepackage{ifthen}  

\newbool{short}
\setboolean{short}{true}

\NewEnviron{shortOnly}{\ifthenelse{\boolean{short}}{\BODY\ }{}\ignorespaces}

\NewEnviron{longOnly}{\ifthenelse{\boolean{short}}{}{\BODY}\ignorespaces}

\NewEnviron{shortAppendix}[1]
    {\ifshort
        \expandafter\global\expandafter\let\csname putShortAppendix#1\endcsname\BODY%
    \else
        \expandafter\newcommand\csname putShortAppendix#1\endcsname{}\BODY%
    \fi}
\newcommand{\putShortAppendix}[1]{\csname putShortAppendix#1\endcsname}

\aclfinalcopy

\newcommand{\xchapter}{\ifthenelse{\boolean{short}}{\section}{\chapter}}
\newcommand{\xsection}{\ifthenelse{\boolean{short}}{\subsection}{\section}}
\newcommand{\xsubsection}{\ifthenelse{\boolean{short}}{\subsubsection}{\subsection}}

\title{Word Equations: Inherently Interpretable Sparse Word Embeddings through Sparse Coding}

\author{Adly Templeton \\
  Williams College \\
  \texttt{adlytempleton@gmail.com}}

\date{}
\begin{document}

\maketitle
\begin{abstract}
Word embeddings are a powerful natural
language processing technique, but they are
extremely difficult to interpret. To enable interpretable NLP models, we create vectors where each dimension is \emph{inherently interpretable}. By inherently interpretable, we mean a system where each dimension is associated with some human-understandable \emph{hint} that can describe the meaning of that dimension. In order to create more interpretable word embeddings, we
transform pretrained dense word embeddings
into sparse embeddings. These new embeddings
are inherently interpretable: each of their
dimensions is created from and represents
a natural language word or specific grammatical
concept. We construct these embeddings
through sparse coding, where each vector in
the basis set is itself a word embedding. Therefore, each dimension of our sparse vectors corresponds to a natural language word. We also show that models trained using these sparse
embeddings can achieve good performance and
are more interpretable in practice, including through human evaluations.
    
\end{abstract}
\xchapter{Introduction}
\label{sec:background}
\textit{Word embeddings} represent each word in a natural language as a vector in a continuous high dimensional space. Many different pretrained embeddings are readily available \cite{mi13,glove,fasttext}, and are used in a range of applications \cite{li18}. This vector representation can be said to encode the meaning of the  word; not only are similar words close together but linear relationships between words are thought to have conceptual meaning. In the famous example, the vector difference between `man' and `woman' is similar to the vector difference between `king' and `queen' \cite{la97,mi13}. This observation suggests that the vector difference between 'woman' and 'man' represents a concept of gender within the vector space, implying that dimensions or linear combinations of dimensions in the vector space are related to human-understandable concepts. However, in practice, interpreting these vector spaces is extremely difficult. This obscures the behavior of any NLP model built on top of word embeddings. 

To enable interpretable NLP models, we create vectors where each dimension is \emph{inherently interpretable}. By inherently interpretable, we mean a system where each dimension is associated with some human-understandable \emph{hint} that can describe the meaning of that dimension. This allows us to directly interpret the coefficients of simple models trained on these vectors. By comparison, most other systems of interpretable word embeddings aim to create dimensions that humans may be able to manually interpret after the fact.

To create our vectors, we represent word embeddings as the sparse linear combination of a basis set of other word embeddings. Our primary contribution is that, instead of learning an optimal basis for our sparse vector space, we draw the columns of the basis from the original set of dense word embeddings. This strategy provides a natural label for each sparse dimension and allows us to represent each natural language word as the linear combination of a small number of other natural language words. This representation is itself a more `interpretable' word embedding. This technique produces representations of words that have interpretable dimensions. We show that these representations are more interpretable and that models trained on these embeddings perform almost as well as models trained on standard dense embeddings. We show how the creation of inherently interpretable vectors can help us understand the behavior and structure of the original word embeddings. 

Recent work has created more interpretable vectors through a variety of methods. However, relatively few approaches create \emph{inherently interpretable} dimensions. Therefore, we believe that our work, which creates inherently interpretable embeddings through a simple novel method can be the basis of future NLP tools where interpretability is crucial.
\nocite{mi13,la97}

As an example, we present one randomly selected embedding from our system. More examples can be found in the appendix.

\begin{align}
&\text{carbon} = 0.79 * \text{nitrogen}\nonumber\\
& -0.38 * \text{CAPITALIZATION} + 0.3 * \text{fossil}\nonumber\\
& -0.21 * \text{POS-NOUN} + 0.16 * \text{POS-ADJ}\nonumber\\
& + 0.14 * \text{C0} -0.14 * \text{PAST-TENSE
}\nonumber\\
& + 0.13 * \text{wood} + 0.11 * \text{global}\nonumber\\
& + 0.1 * \text{atoms} -0.095 * \text{POS-ADV}\nonumber\\
& + 0.092 * \text{aluminum}\nonumber\\
& -0.078 * \text{PLURAL-NOUN}\nonumber\\
& + 0.073 * \text{greenhouse}\nonumber\\
 & -0.072 * \text{POS-PROPN}\nonumber\\
& -0.048 * \text{POS-VERB} + 0.046 * \text{forestry}\nonumber\\
& + 0.03 * \text{PARTICIPLE
} + 0.017 * \text{sink}\nonumber\\
& + 0.012 * \text{POS-NUM}\nonumber
\end{align}

\xchapter{Previous Work}
\label{sec:word_intrusion}
Park et al. \cite{pa17} find a more interpretable rotation of word embeddings using techniques associated with factor analysis. Other work \cite{du19,ro16} rotates dense vectors using different methods. 

Koc et al.  \cite{ko18} tie concepts to dimensions in a more direct way. They select a concept for each dense dimension and identify words that are associated with these concepts\begin{shortOnly}. A penalty term pushes coefficients for these words towards the fixed values.
\end{shortOnly}

Other work has focused on interpretability through sparsity. Subramian et al. \cite{su18} created more interpretable embeddings by passing pretrained dense embeddings through a sparse autoencoder. 

Panigrahi et al. \cite{pa19} proposed Word2Sense, a generative approach that models each dimension as a `sense' and word embeddings as a sparse probability distribution over the senses. 

The mathematical technique we use in this paper, \emph{Sparse coding}, which is defined as the representation of vectors as the sparse linear combination of an overcomplete basis, is a well-studied optimization problem \cite{co11,ho02,le07}. Previous work  \cite{co11} has also shown that basis vectors can be efficiently selected from the set that is being encoded.

Faruqui et al. \cite{fa15} used non-negative sparse coding to recode dense word embeddings into more interpretable sparse vectors while learning a basis. However, because they create their basis through direct optimization, the basis vectors (and, consequently, the dimensions in their transformed sparse space) do not have any inherent interpretation and must be manually interpreted.

Zhang et al. \cite{zh19b} also used non-negative sparse coding to learn a set of \textit{word factors} to recode word2vec embeddings. The basis vectors created in this way are highly redundant, so they then use spectral clustering to remove near-duplicate factors. Then, they are able to manually infer reasonable post hoc interpretations for most of the factors.

Concurrently with our work, Mathew et al. create an inherently interpretable subspace from pairs of antonyms. They then project embeddings into that subspace, producing lower-dimensional dense vectors \cite{polar}. 

\xchapter{Model}
\label{sec:methodology}
Our work uses \emph{sparse coding} to transform a set of word embeddings from a dense and uninterpretable space into a sparse and interpretable space. Let $v_D$ represent a dense word embedding, and let $\mathcal{B}$ represent a matrix with basis vectors along the columns.  $\mathcal{B}$ has size $(n_S, n_N)$ where $n_d$ is the dimensionality of the dense vectors and $n_S$ is the dimensionality of the sparse vectors. We achieve sparse coding using regularized regression, inducing sparsity using the $L_1$ norm.
Formally, this corresponds to finding the sparse vector $v_S$ that minimizes the following objective function
\begin{equation}
\label{eq:lasso}
\argmin{v_S} ~||v_D - v_S\mathcal{B}||_2^2 + \alpha||v_S||_1
\end{equation}
\noindent $\alpha$ is a hyperparameter that controls the level of sparsity. The first term in Equation \ref{eq:lasso} ensures the sparse vector corresponds to a vector in the dense space that is similar to the original vector. The second term is a sparsity-inducing penalty.

Note that by `basis` we mean a set of vectors in the dense space, each one corresponding to a dimension in the transformed, sparse, space. Out of necessity, these vectors are \textit{overcomplete} (there are more dimensions than vectors) and so they do not form a basis according to the traditional definition.

Previous work using sparse coding to create interpretable word embeddings has considered the basis $\mathcal{B}$ to be part of the optimization problem \cite{fa15, zh19b}. Our primary contribution is that, instead of learning an optimal basis, we draw the columns of the basis from the original set of dense word embeddings. This strategy provides a natural label for each sparse dimension.

\xsection{Grammatical Basis}

We can roughly divide the `meaning' carried by a word embedding into \emph{grammatical} and \emph{non-grammatical} properties. Here we use `grammatical properties' to mean properties that describe how that word fits into the grammar of the language, such as its part-of-speech, tense, or number. We use `non-grammatical properties' to mean all other aspects of the meaning of a word. For instance, we expect the embedding for the word `swimming' to include a grammatical component representing that this word is a present-tense participle and a non-grammatical component that represents the meaning `to swim'. Of course, this deconstruction is imperfect. Nevertheless, this approach provides a useful insight towards decomposing the meaning of a word embedding.

Preliminary experiments showed that, without special consideration, grammatical properties would be captured in an unintuitive way. The grammatical components could not be easily isolated to one subset of the nonzero dimensions. Ideally, the grammatical information would be captured in a small number of interpretable dimensions. Instead, each basis vector would capture part of the grammatical component and part of the semantic component. This duality creates difficulty when interpreting our representations.

To address this, we construct a small number of \emph{grammatical basis vectors} and add them to the basis set. For instance, we construct a `POS-NOUN' vector by taking the mean of all word embeddings corresponding to nouns. For this work, we use a set of 11 grammatical basis vectors, though the number and the construction of these are arbitrary. A description of the grammatical basis vectors is in the appendix.

\begin{shortAppendix}{gramBasis}
Our approach makes use of four types of grammatical basis vectors:

\begin{enumerate}
    \item We use the first principal component of the embeddings of the 30,000 most frequent words. Previous work on word embedding has referred to this as the \emph{common discourse vector}, or $c_0$, and has shown that this vector encodes words that appear commonly in all contexts, such as `the'.
    \item We take the mean of all vectors of capitalized words and use this as a grammatical basis vector to represent capitalization.
    \item For a variety of parts-of-speech, we use the mean vectors for words with that part-of-speech (POS). Specifically, we encode a vector for each of the following: nouns, verbs, adjectives, adverbs, and numbers.
    \item We create mean vector differences for the following grammatical concepts: the relationship between singular and plural nouns, the relationship between present-tense verbs and their present participle form, and the relationship between present-tense verbs and their past-tense forms. For each of these relationships, we manually collect approximately 50 example word pairs that fit that relationship. We manually filter for word pairs where either the grammatical relationship does not change the form of the word (i.e., `deer') or for word pairs where the grammatical change is likely to produce a more complicated change in meaning (i.e., `math' and `maths'). We average the differences between pairs of each relationship type and use it as the vector for that relationship.
\end{enumerate}

The choice and construction of these grammatical basis vectors is highly arbitrary, and different grammatical basis vectors could easily be used in different applications or in follow up work.
\end{shortAppendix}

Next, we make the grammatical basis vectors orthogonal using the Gram-Schmidt process. Finally, we subtract the projection along the grammatical basis vectors from all other (`non-grammatical') basis vectors we use and renormalize them. This procedure separates the grammatical meaning from our non-grammatical basis vectors, ensuring that non-grammatical bases are not also coding for grammatical concepts.

Note that we only perform this orthogonalization with respect to a very small number of grammatical basis vectors. We find that this procedure does not remove more than 50\% of the length of any individual vector and 50\% of vectors have less than 20\% of their length removed.

When encoding a dense vector, instead of finding the grammatical coefficients using sparse coding, we set each grammatical coefficient to the projection along the corresponding grammatical basis vector, which is equal to the dot product similarity between the original vector and the grammatical basis vector. Because the grammatical basis is orthogonal, we can do this for every grammatical basis vector simultaneously. This residual is then transformed using Equation \ref{eq:lasso}. 

Note that, although we do require hand-crafted features to create the grammatical basis vectors, our system does not use hand-crafted features in the representation of new words. Once the grammatical feature vectors are defined, words can be represented in our sparse space using no more information than their \textit{fasttext} dense vectors.

\xsection{Basis Selection}

We cannot practically use all words as our basis set, so we have to select a subset. First, we start with the 30,000 most frequent words. We filter out any words that are capitalized or that are not in a standard English vocabulary (using the vocabulary of the spaCy \emph{en\_core\_web\_sm} model). Next, we filter out any words that are not nouns, verbs, or adjectives. This process removes many basis vectors that may be hard to interpret. This gives us approximately 11,000 remaining potential basis words. From these, we will select 3,000 words to use in the final basis. 

We use an iterative algorithm that takes, at each step,  the potential basis vector with the highest mean cosine similarity to all other vectors. To encourage diversity, this mean is weighted by the lowest cosine dissimilarity that each vector has with any already-selected basis vector. Formally, at each step, we grow the set of basis vectors $B$ by adding the potential basis vector $x$ from the set of unchosen potential basis vectors $\mathcal{F}\setminus B$ that satisfies

$$\argmax{x \in \mathcal{F}\setminus B} \sum_{v \in V_D} (x \cdot v) \max_{b \in B} (1 - b \cdot v)$$

\noindent Where $V_D$ is the set of dense vectors for the 30,000 most frequent words.

Note that, despite our use of the word `basis', this is not a basis in the traditional sense; the set of basis vectors are not linearly independent, and there are more basis vectors than dimensions in the original space. However, because of the L1 penalty term, our objective function still allows for optimal decompositions.  

\xsection{End-to-end Process}
In order to find the sparse vector representation, we follow the following process, combining the above elements.

\begin{enumerate}
    \item Find the dense vector representation of the word.
    \item Compute the projection along each vector orthogonal grammatical basis. Store these projections as the first part of the resulting vector. Subtract the projection along this basis before moving on to the next step.
    \item Optimize Eq. \ref{eq:lasso} using the FISTA algorithm \cite{fista}. Store the learned sparse vector as the second part of the resulting vector. 
\end{enumerate}

\xchapter{Results}
We will evaluate this model in multiple ways. In particular, we care about two  contradictory properties of our transformed vector space. First, we want our vector space to be useful in downstream machine learning applications. We expect that, in most applications, increased interpretability comes with some performance cost. Therefore, we care about the performance loss when moving from dense vectors to our sparse vectors.

The other goal is that our sparse vectors should be interpretable. It is much harder to articulate exactly what interpretability is or how we can measure it. Metrics such as the Word Intrusion Task (Section\ref{sec:word_intrusion}) can act as a useful proxy for interpretability, and we use it as our primary quantitative measure of interpretability. But part of interpretability is, by definition, subjective and any metric is imperfect. 
\xsection{implementation}
We use the FastText \cite{fasttext} pretrained 300 dimensional English vectors (without subword information) trained on Wikipedia 2017, UMBC webbase corpus and statmt.org news dataset as the dense vectors that we input into our models. Unless otherwise mentioned, we only consider the 30,000 most frequent words, for computational reasons. We normalize all vectors to have mean 0 and unit length. After learning sparse vectors, we normalize each sparse vector so that it corresponds to a dense vector of unit length. When comparing with the original dense vectors (FastText \cite{fasttext}), we subtract the mean of all vectors, to match our preprocessing.

In practice, the sparse penalty term will only push coefficients very close to 0. We clamp any coefficient with a magnitude of less than .001 to 0. We found this threshold by taking the lowest cutoff that does not introduce significant irregularities into the tradeoff curves in Section \ref{sec:tradeoff}.

We solve the regularized optimization problems using the FISTA algorithm \cite{fista}, as implemented in the Python Lightning package \cite{lightning_2016}, using default hyperparameters. FISTA is an optimization algorithm that can efficiently solve sparse coding problems. We use the spaCy library \cite{spacy2}  to check for out of vocabulary words and perform part-of-speech tagging. We use the numpy \cite{ol06}, CuPy \cite{cu17}, and Scikit learn \cite{pe11} libraries for various linear algebra implementations. We use the open-source Gensim library \cite{gensim} to manipulate word embeddings. \begin{shortOnly}\ For the word analogy task evaluation, we use the \emph{3CosAdd} method, as implemented by Gensim. Models processed 30,000 words within a few hours, running across 32 2.5 GHz processors with no GPU.\end{shortOnly}
\nocite{fista}
\nocite{spacy2}
\nocite{ol06}
\nocite{cu17}
\nocite{pe11}

\xsection{Comparison with Previous Work}
\label{sec:Faruqui_comp}
To compare our work against other sparse coding approaches, we will often reference the vectors created by Faruqui et al. \cite{fa15}. That work generates more interpretable vectors using sparse coding but without inherently interpretable dimensions. \begin{shortAppendix}{implementation_Faruqui} 

To compare to the sparse coding approach of Faruqui et al., we use their publicly available implementation with the following settings: We use the same input vectors without preprocessing, a dimensionality of 3000, $L_2$ regularization penalty $\tau = 10^{-5}$, as suggested in their paper, and various $L_1$ regularization penalties ($\lambda$).\end{shortAppendix}

\xsection{Reconstruction Error and Sparsity}
\label{sec:tradeoff}

Note that, because of the penalty term in Equation \ref{eq:lasso}, $V_S\mathcal{B}$ (the \emph{reconstructed vectors}) are not exactly equal to the original dense vectors $V_D$. Therefore, we expect a tradeoff between sparsity and this difference (which we call \emph{reconstruction error)}.

\begin{figure}
    \centering
    \includegraphics[scale=.4]{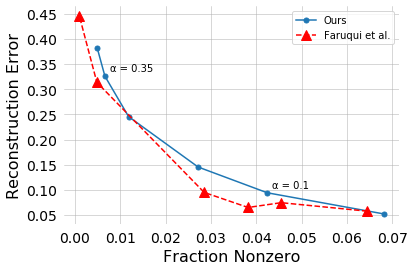}
    \caption{The tradeoff curve between sparsity and reconstruction error. The dashed line shows the tradeoff curve achieved by Faruqui et al. using sparse coding without inherently interpretable dimensions (Section \ref{sec:Faruqui_comp}).}
    \label{fig:sparsity_tradeoff}
\end{figure}

This tradeoff curve is displayed in Figure \ref{fig:sparsity_tradeoff}. Despite the additional constraints of an inherently interpretable system, we suffer only a minor increase in reconstruction error compared to traditional sparse coding. This \emph{reconstruction error} is the primary drawback of our system; reconstruction error adds a small amount of noise to every model built on top of our sparse vectors. For the remainder of this work, unless otherwise mentioned, we will consider the vectors made with $\alpha = 0.35$. These vectors have, on average, 20 nonzero entries.

\xsection{Analogy Task}
\label{sec:results_analogy}
In the word2vec vector space, famously, the vector for `king' plus the vector for `woman' minus the vector for `man' is close to the vector for `queen'. Analogy tasks quantitatively test these properties. The task consists of analogies of the form \emph{$A$ is to $A'$ as $B$ is to $B'$}. The vector space is evaluated on its ability to correctly determine the value of $B'$.

The performance of our vector space at this task is displayed in Table \ref{tab:analogy}. Our model performs poorly on this task. This degradation comes from two sources. First, the drop from the original vectors to the reconstructed vectors that is due to reconstruction error. Second, an additional degradation is caused by the transformation from dense vectors to sparse vectors, especially with cosine similarity.

\begin{table}
\centering
\small
\begin{tabular}{|c|c|ccc|}
\hline
                             & Nonzero  &  Total & Gram. & Sem. \\ \hline
FastText  & 300 &0.88 & 0.85 & 0.94\\
Faruqui $\lambda = 0.75$   &    136 &0.65 & 0.60 & 0.73\\ \hline
Ours $\alpha = 0.1$ & 127    &  0.50 & 0.46 & 0.59\\
Recons. $\alpha = 0.1$ &  300       & 0.83 & 0.80 & 0.88\\ \hline
Ours $\alpha = 0.35$ &            20                   & 0.20 & 0.22 & 0.15\\
Recons. $\alpha = 0.35$&       300   &0.33 & 0.38 & 0.25\\  
\hline
\end{tabular}
\caption[Analogy evaluation results]{Accuracy on the word2vec analogy evaluation set for various vector spaces. The first column shows the average number of nonzero entries in each sparse vector. Accuracy is also broken down by non-grammatical and grammatical categories. 'Recons' denotes the performance of the reconstructed dense vectors. All results are on a 50\% held-out test set.}
\label{tab:analogy}
\end{table}

\xsection{Classification}
Next, we demonstrate that our model can be used to build interpretable machine learning systems. To this end, we train classifiers using our word embeddings as input. We demonstrate that these classifiers are not only effective but also interpretable.


We evaluate our vectors on two datasets, the IMDB sentiment analysis dataset \cite{imdbmovies} and the TREC question classification dataset \cite{trec}. For both of these datasets, we use a logistic regression model and a bag of words representation.

\xsubsection{IMDB Sentiment Analysis Dataset}

The IMDB movie review dataset consists of 50,000 passages taken from IMDB movie reviews, evenly split between positive and negative reviews. The task is to determine the sentiment of each passage \cite{imdbmovies}.

We train classifiers using various word embedding spaces as inputs. While we could train deep neural modesl on these vector spaces, neural models do not directly produce interpretable coefficients, and therefore we provide a demonstration on simple logistic regression models. The results are presented in Table \ref{tab:imdb_results}. Our vector spaces demonstrate improvement over the original dense vectors (FastText \cite{fasttext}), as well as the traditional sparse coding approach of Faruqui et al. This result holds despite a slight decrease in performance caused by the reconstruction error (as demonstrated by the low performance with reconstructed vectors).

\begin{table}[]
\centering
\begin{tabular}{|l|l|l|}
\hline
                              & IMDB & TREC\\ \hline
FastText     &  85.35 & 84.2\\
Faruqui $\lambda = .75$     & 85.54 & 84.4\\ \hline
Ours $\alpha = 0.1$ &  \textbf{87.51}  & \textbf{86.2}     \\
Ours Recons. $\alpha = 0.1$  &  85.08  & 81.4 \\ \hline
Ours $\alpha = 0.35$ &  86.46 & 84.0\\
Ours Recons. $\alpha = 0.35$&    83.00 & 75.8  \\
\hline
\end{tabular}
\caption[Results]{Accuracy on the IMDB sentiment analysis dataset and the TREC question classification dataset. We use a logistic regression classifier, which uses as input a bag-of-words sum of various word embeddings.}
\label{tab:imdb_results}
\end{table}

We can directly interpret our classifier's coefficients. Here, we present the most significant coefficients ($\alpha = 0.1$) \footnote{These weights are real-values and truncated for space. Note that the weights are very large because they correspond to sparse low-magnitude features.}:
\begingroup
\addtolength{\jot}{.1em}
\begin{align*}
     \ln &\frac{P(\text{positive})}{1 - P(\text{positive})} = -157 \cdot \text{dreadful}\\&
     - 153 \cdot \text{horrible} + 150 \cdot \text{fabulous} -140 \cdot \text{dull} \\&
     -132 \cdot \text{dreary} -107 \cdot \text{worsen}
     \\&-105 \cdot \text{ridiculous} + ... \\&
\end{align*}
\endgroup

\noindent Note that these are not coefficients on the frequencies of individual words. Instead, these are coefficients on vectors in the basis set. We can consider them to be coefficients on concepts, which are labeled by the displayed words. The coefficients make sense: positive concepts have positive coefficients, while negative concepts have negative coefficients. This pattern continues for much longer than displayed above, and we have omitted other terms for space reasons. The first term to not fall into this clear interpretation is the 24th-most significant:\begin{shortOnly}\ $~... + 74 \cdot \text{shall} + ..`$ \end{shortOnly}

\noindent At first, this term appears nonsensical. Looking more closely at this dimension can reveal more about our system.  The top five words in the dimension represented by `shall' are the following: `henceforth', `herein', `hereafter', `thereof', `hereby'. We can see here how both our vector space and our regression model pick up on tone. This dimension appears to correspond to a formal and somewhat archaic tone, which is likely not found in a negative internet comment.

\xsubsection{TREC Question Classification Dataset}

Our next classification task is more complex. The TREC question classification dataset consists of 6,000 questions that are divided into 6 categories based on the expected answer: abbreviations, descriptions, entities, humans, locations, and numeric.

Accuracy for various vector spaces is presented in Table \ref{tab:imdb_results}. Again, our model does better than the unmodified input vectors we start with, despite some loss from the reconstruction error. Both results suggest that our vector spaces are efficient in regression-based settings, though the performance at the word-analogy task suffers a serious degradation. It is likely that different qualities are needed for these different tasks. 
The exact-match evaluation of the word analogy task severely punishes even slight noise in the vector space, and cosine similarities are noisy in sparse vectors.

Once again, we directly interpret the coefficients learned by logistic regression. For space, we display the most significant terms for the HUM category. Questions in this category expect the name of a human as the answer:

\begin{figure*}[h!]
    \centering
    \fbox{\includegraphics[scale=1]{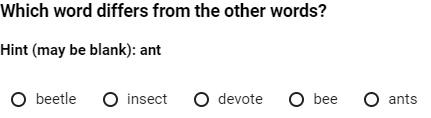}}
    \caption[Word intrusion interface example]{An example of the user interface given to annotators. The following instructions were given to the annotators: `You will be presented with a group of 5 words.
    Four of these words are similar in some way and the other one is not.
    Pick out the word which is dissimilar.
    You may be provided with a hint about how the words are similar.'}
    \label{fig:mturk_gui}
\end{figure*}

\begingroup
\addtolength{\jot}{.1em}
\begin{align*}
    \ln &\frac{P(\text{HUM})}{1 - P(\text{HUM})} = - 77 \cdot \text{wonder}\\&
    + 66 \cdot \text{organizations}+ 54 \cdot \text{companies}\\&+ 51 \cdot \text{poet}+ 49 \cdot \text{songwriter} \\& + 48 \cdot \text{identities}+ 42 \cdot \text{fan}- 42 \cdot \text{movie} \\&- 39 \cdot \text{resulting}+ 36 \cdot \text{university}\\&  - 36 \cdot \text{diseases}+ 36 \cdot \text{successive}+\\& 35 \cdot \text{consist} + 35  \cdot \text{cabinets} + ...
\end{align*}
\endgroup

Some of these coefficients, such as `songwriter' or `identities' are intuitive and reveal interesting behavior of the classifier. Others, such as `wonder', are not. Manual inspection reveals that `wonder' is used to represent words such as `How' or `why' but not `who', though this behavior is likely noise.

\xsubsection{Word Intrusion Task}
To quantitatively measure interpretability, we use human experiments. In particular, we use the word intrusion task \cite{ch09}. In this task, humans are presented with five words, four of which are associated highly with a particular dimension. Participants are asked to choose the word that does not belong. We use our vectors both with and without providing the label of the dimension as a `hint'.

We use the following procedure for generating questions. First, we filter candidate words, starting with the 20,000 most frequent words and filtering out words that are not lowercase, words that are not made up of only ASCII alphabetic characters, and words with only one letter. Then we randomly select a dimension. We pick the 4 highest words along that dimension, and one word randomly selected from the bottom 50\% of words in that dimension, then randomize the order.  Each example is presented to three different Mechanical Turk annotators. An example of the interface presented to annotators is seen in Figure \ref{fig:mturk_gui}.

\begin{shortOnly}

\begin{table}[]
\centering
\begin{tabular}{|l|ll|}
\hline
                      & \textbf{Accuracy} & \textbf{CI}     \\ \hline
\textbf{FastText}     & 0.31    & {[}0.27,0.34{]} \\
\textbf{Faruqui}      & 0.77   & {[}0.74,0.80{]} \\
\textbf{Ours}         & 0.80     & {[}0.77,0.83{]} \\
\textbf{Ours (with Hints)} & 0.84  & {[}0.81,0.86{]} \\ \hline
\end{tabular}

\caption[Word intrusion results]{Results on the word intrusion task. 95\% normal confidence intervals are displayed.}
\label{tab:intrusion_results}
\end{table}
\end{shortOnly}

\noindent The results of the word intrusion task are presented in Table \ref{tab:intrusion_results}. When hints are provided, we see a statistically significant improvement in accuracy between our vectors and the sparse coding baseline (p = .00055). In addition, using hints produces a statistically significant improvement (p = .040), validating our motivation for inherently interpretable dimensions. Of course, any quantitative metric of interpretability is imperfect. To qualitatively assess interpretability, randomly selected vectors are presented in the appendix.

\xsection{Summary}

Our method still has some serious drawbacks. Sparse coding, by its nature, introduces a substantial amount of noise in the form of reconstruction error and sparse coding has the potential to assign very different sparse vectors to similar dense vectors. We hope that future work will produce sparse embeddings that are interpretable by construction without some of the shortcomings of our work.

\xsection{Conclusions and Future Work}
In this work, we presented a method to create word embeddings that are interpretable by construction. Each dimension of these embeddings corresponds precisely to a natural language word. These embeddings can be presented in a human readable form, and we have shown that most of these representations are intuitive. We have also shown that these embeddings can be used to produce an extremely interpretable classification model that still delivers performance comparable to or better than a classification model based on the original embeddings.

Unlike most previous work on interpretable word embeddings, our method does not require humans to interpret and label each dimension. We have previously seen how this feature allows us to easily create interpretable classification models. It also allows us to gain a deeper understanding of the original dense vector space. Previous approaches may have obscured nuanced or hard to interpret behavior. In particular, a human manually interpreting a dimension may not appreciate subtle behavior of the system. Several sections of this work, which have manually examined individual word representations in our system, have revealed the nuanced behavior that our system demonstrates.

Our method still has some serious drawbacks. While we have examined a number of these flaws, many are tied closely to the sparse coding method we have chosen to use. Sparse coding, by its nature, introduces a substantial amount of noise in the form of reconstruction error. In addition to the reconstruction error, sparse coding has the potential to assign very different sparse vectors to similar dense vectors. We hope that future work will produce sparse embeddings that are interpretable by construction without some of the shortcomings of our work.

Much of the promise of sparse coding methods remains to be proved. In particular, we believe it will be fruitful to study the representation of syntactic concepts. We have seen that our attempts to disentangle syntactic concepts from our semantic basis vectors were not entirely successful. We would also like to better understand how these methods are applicable in deep learning models. 

There is still a large amount of analytical work left to be done on evaluation. The word intrusion task, while an effective quantitative method, does not offer a complete view of interpretability. Part of this problem is that we do not have any way to quantify interpretability where it is most useful: when building downstream classification models. More fundamentally, we do not have any underlying framework for understanding what it means for a word embedding to be interpretable.

We believe that interpretable word embeddings have great potential for helping us understand and interpret models in a wide range of NLP tasks.

\section*{Acknowledgements}
Thanks to Duane Bailey for his extensive support and advice, and for advising the thesis on which this paper is based.. Thanks to Andrea Danyluk for her guidance as the second reader of that thesis. 

\clearpage
\bibliographystyle{acl_natbib}
\bibliography{emnlp2020}

\begin{thebibliography}{27}
\expandafter\ifx\csname natexlab\endcsname\relax\def\natexlab#1{#1}\fi

\bibitem[{Blondel and Pedregosa(2016)}]{lightning_2016}
Mathieu Blondel and Fabian Pedregosa. 2016.
\newblock \href {https://doi.org/10.5281/zenodo.200504} {{Lightning:
  large-scale linear classification, regression and ranking in Python}}.

\bibitem[{Bojanowski et~al.(2017)Bojanowski, Grave, Joulin, and
  Mikolov}]{fasttext}
Piotr Bojanowski, Edouard Grave, Armand Joulin, and Tomas Mikolov. 2017.
\newblock Enriching word vectors with subword information.
\newblock \emph{Transactions of the Association for Computational Linguistics},
  5:135--146.

\bibitem[{Chalasani et~al.(2013)Chalasani, Principe, and Ramakrishnan}]{fista}
Rakesh Chalasani, Jose~C Principe, and Naveen Ramakrishnan. 2013.
\newblock A fast proximal method for convolutional sparse coding.
\newblock In \emph{The 2013 International Joint Conference on Neural Networks
  (IJCNN)}, pages 1--5. IEEE.

\bibitem[{Chang et~al.(2009)Chang, Gerrish, Wang, Boyd-Graber, and Blei}]{ch09}
Jonathan Chang, Sean Gerrish, Chong Wang, Jordan~L. Boyd-Graber, and David~M.
  Blei. 2009.
\newblock Reading tea leaves: How humans interpret topic models.
\newblock In \emph{Advances in Neural Information Processing Systems}, pages
  288--296.

\bibitem[{Coates and Ng(2011)}]{co11}
Adam Coates and Andrew~Y. Ng. 2011.
\newblock The importance of encoding versus training with sparse coding and
  vector quantization.
\newblock In \emph{Proceedings of the 28th International Conference on Machine
  Learning (ICML-11)}, pages 921--928.

\bibitem[{Dufter and Sch{\"u}tze(2019)}]{du19}
Philipp Dufter and Hinrich Sch{\"u}tze. 2019.
\newblock Analytical methods for interpretable ultradense word embeddings.
\newblock \emph{arXiv preprint arXiv:1904.08654}.

\bibitem[{Faruqui et~al.(2015)Faruqui, Tsvetkov, Yogatama, Dyer, and
  Smith}]{fa15}
Manaal Faruqui, Yulia Tsvetkov, Dani Yogatama, Chris Dyer, and Noah~A Smith.
  2015.
\newblock Sparse overcomplete word vector representations.
\newblock In \emph{Proceedings of the 53rd Annual Meeting of the Association
  for Computational Linguistics and the 7th International Joint Conference on
  Natural Language Processing (Volume 1: Long Papers)}, pages 1491--1500.

\bibitem[{Honnibal and Montani(2017)}]{spacy2}
Matthew Honnibal and Ines Montani. 2017.
\newblock {spaCy 2}: Natural language understanding with {B}loom embeddings,
  convolutional neural networks and incremental parsing.

\bibitem[{Hoyer(2002)}]{ho02}
Patrik~O. Hoyer. 2002.
\newblock Non-negative sparse coding.
\newblock In \emph{Proceedings of the 12th IEEE Workshop on Neural Networks for
  Signal Processing}, pages 557--565. IEEE.

\bibitem[{Landauer and Dumais(1997)}]{la97}
Thomas~K. Landauer and Susan~T. Dumais. 1997.
\newblock A solution to {P}lato's problem: The latent semantic analysis theory
  of acquisition, induction, and representation of knowledge.
\newblock \emph{Psychological Review}, 104(2):211.

\bibitem[{Lee et~al.(2007)Lee, Battle, Raina, and Ng}]{le07}
Honglak Lee, Alexis Battle, Rajat Raina, and Andrew~Y. Ng. 2007.
\newblock Efficient sparse coding algorithms.
\newblock In \emph{Advances in Neural Information Processing Systems}, pages
  801--808.

\bibitem[{Li and Roth(2002)}]{trec}
Xin Li and Dan Roth. 2002.
\newblock Learning question classifiers.
\newblock In \emph{Proceedings of the 19th International Conference on
  Computational Linguistics-Volume 1}, pages 1--7. Association for
  Computational Linguistics.

\bibitem[{Li and Yang(2018)}]{li18}
Yang Li and Tao Yang. 2018.
\newblock Word embedding for understanding natural language: a survey.
\newblock In \emph{Guide to Big Data Applications}, pages 83--104. Springer.

\bibitem[{Maas et~al.(2011)Maas, Daly, Pham, Huang, Ng, and Potts}]{imdbmovies}
Andrew~L. Maas, Raymond~E. Daly, Peter~T. Pham, Dan Huang, Andrew~Y. Ng, and
  Christopher Potts. 2011.
\newblock \href {http://www.aclweb.org/anthology/P11-1015} {Learning word
  vectors for sentiment analysis}.
\newblock In \emph{Proceedings of the 49th Annual Meeting of the Association
  for Computational Linguistics: Human Language Technologies}, pages 142--150,
  Portland, Oregon, USA. Association for Computational Linguistics.

\bibitem[{Mathew et~al.(2020)Mathew, Sikdar, Lemmerich, and Strohmaier}]{polar}
Binny Mathew, Sandipan Sikdar, Florian Lemmerich, and Markus Strohmaier. 2020.
\newblock The polar framework: Polar opposites enable interpretability of
  pre-trained word embeddings.
\newblock \emph{arXiv preprint arXiv:2001.09876}.

\bibitem[{Mikolov et~al.(2013)Mikolov, Sutskever, Chen, Corrado, and
  Dean}]{mi13}
Tomas Mikolov, Ilya Sutskever, Kai Chen, Greg~S Corrado, and Jeff Dean. 2013.
\newblock Distributed representations of words and phrases and their
  compositionality.
\newblock In \emph{Advances in Neural Information Processing Systems}, pages
  3111--3119.

\bibitem[{Okuta et~al.(2017)Okuta, Unno, Nishino, Hido, and Loomis}]{cu17}
Ryosuke Okuta, Yuya Unno, Daisuke Nishino, Shohei Hido, and Crissman Loomis.
  2017.
\newblock \href {http://learningsys.org/nips17/assets/papers/paper_16.pdf}
  {{C}u{P}y: A {N}um{P}y-compatible library for nvidia gpu calculations}.
\newblock In \emph{Proceedings of Workshop on Machine Learning Systems
  (LearningSys) in The Thirty-first Annual Conference on Neural Information
  Processing Systems (NIPS)}.

\bibitem[{Oliphant(2006)}]{ol06}
Travis~E. Oliphant. 2006.
\newblock \emph{A guide to NumPy}, volume~1.
\newblock Trelgol Publishing USA.

\bibitem[{Panigrahi et~al.(2019)Panigrahi, Simhadri, and Bhattacharyya}]{pa19}
Abhishek Panigrahi, Harsha~Vardhan Simhadri, and Chiranjib Bhattacharyya. 2019.
\newblock Word2sense: Sparse interpretable word embeddings.
\newblock In \emph{Proceedings of the 57th Annual Meeting of the Association
  for Computational Linguistics}, pages 5692--5705.

\bibitem[{Park et~al.(2017)Park, Bak, and Oh}]{pa17}
Sungjoon Park, JinYeong Bak, and Alice Oh. 2017.
\newblock Rotated word vector representations and their interpretability.
\newblock In \emph{Proceedings of the 2017 Conference on Empirical Methods in
  Natural Language Processing}, pages 401--411.

\bibitem[{Pedregosa et~al.(2011)Pedregosa, Varoquaux, Gramfort, Michel,
  Thirion, Grisel, Blondel, Prettenhofer, Weiss, Dubourg et~al.}]{pe11}
Fabian Pedregosa, Ga{\"e}l Varoquaux, Alexandre Gramfort, Vincent Michel,
  Bertrand Thirion, Olivier Grisel, Mathieu Blondel, Peter Prettenhofer, Ron
  Weiss, Vincent Dubourg, et~al. 2011.
\newblock Scikit-learn: Machine learning in {P}ython.
\newblock \emph{Journal of Machine Learning Research}, 12(Oct):2825--2830.

\bibitem[{Pennington et~al.(2014)Pennington, Socher, and Manning}]{glove}
Jeffrey Pennington, Richard Socher, and Christopher~D Manning. 2014.
\newblock Glove: Global vectors for word representation.
\newblock In \emph{Proceedings of the 2014 Conference on Empirical Methods in
  Natural Language Processing (EMNLP)}, pages 1532--1543.

\bibitem[{Rehurek and Sojka(2010)}]{gensim}
Radim Rehurek and Petr Sojka. 2010.
\newblock Software framework for topic modelling with large corpora.
\newblock In \emph{In Proceedings of the LREC 2010 Workshop on New Challenges
  for NLP Frameworks}. Citeseer.

\bibitem[{Rothe and Sch{\"u}tze(2016)}]{ro16}
Sascha Rothe and Hinrich Sch{\"u}tze. 2016.
\newblock Word embedding calculus in meaningful ultradense subspaces.
\newblock In \emph{Proceedings of the 54th Annual Meeting of the Association
  for Computational Linguistics (Volume 2: Short Papers)}, pages 512--517.

\bibitem[{{\c{S}}enel et~al.(2020){\c{S}}enel, Utlu, {\c{S}}ahinu{\c{c}},
  Ozaktas, and Ko{\c{c}}}]{ko18}
L{\"u}tfi~Kerem {\c{S}}enel, Ihsan Utlu, Furkan {\c{S}}ahinu{\c{c}}, Haldun~M
  Ozaktas, and Aykut Ko{\c{c}}. 2020.
\newblock Imparting interpretability to word embeddings while preserving
  semantic structure.
\newblock \emph{Natural Language Engineering}, pages 1--26.

\bibitem[{Subramanian et~al.(2018)Subramanian, Pruthi, Jhamtani,
  Berg-Kirkpatrick, and Hovy}]{su18}
Anant Subramanian, Danish Pruthi, Harsh Jhamtani, Taylor Berg-Kirkpatrick, and
  Eduard Hovy. 2018.
\newblock Spine: Sparse interpretable neural embeddings.
\newblock In \emph{Thirty-Second AAAI Conference on Artificial Intelligence}.

\bibitem[{Zhang et~al.(2019)Zhang, Chen, Cheung, and Olshausen}]{zh19b}
Juexiao Zhang, Yubei Chen, Brian Cheung, and Bruno~A. Olshausen. 2019.
\newblock Word embedding visualization via dictionary learning.
\newblock \emph{arXiv preprint arXiv:1910.03833}.

\end{thebibliography}
\clearpage
\xchapter{Appendix}
\appendix
\section{Gramtacic Basis Descriptions}
\putShortAppendix{gramBasis}
\section{Implementation}
\putShortAppendix{implementation}
\subsection{Comparison to Faruqui et al.}
\putShortAppendix{implementation_Faruqui}
\subsection{Word Intrusion Task Implementation}
\putShortAppendix{Word Intrusion}
\section{Randomly Selected Word Representations}
We randomly select 25 words and display their complete sparse vector representations here:

\begin{align}
&\text{carbon} = 0.79 * \text{nitrogen}\nonumber\\
& -0.38 * \text{CAPITALIZATION} + 0.3 * \text{fossil}\nonumber\\
& -0.21 * \text{POS-NOUN} + 0.16 * \text{POS-ADJ}\nonumber\\
& + 0.14 * \text{C0} -0.14 * \text{PAST-TENSE
}\nonumber\\
& + 0.13 * \text{wood} + 0.11 * \text{global}\nonumber\\
& + 0.1 * \text{atoms} -0.095 * \text{POS-ADV}\nonumber\\
& + 0.092 * \text{aluminum}\nonumber\\
& -0.078 * \text{PLURAL-NOUN}\nonumber\\
& + 0.073 * \text{greenhouse}\nonumber\\
 & -0.072 * \text{POS-PROPN}\nonumber\\
& -0.048 * \text{POS-VERB} + 0.046 * \text{forestry}\nonumber\\
& + 0.03 * \text{PARTICIPLE
} + 0.017 * \text{sink}\nonumber\\
& + 0.012 * \text{POS-NUM}\nonumber
\end{align}
\begin{align}
&\text{reefs} = 0.68 * \text{islands}\nonumber\\
& -0.66 * \text{CAPITALIZATION} + 0.4 * \text{C0}\nonumber\\
& + 0.35 * \text{PLURAL-NOUN
}\nonumber\\ &  + 0.28 * \text{POS-VERB}\nonumber\\
& + 0.25 * \text{rocks}\nonumber\\
& + 0.19 * \text{dredging}\nonumber\\
& + 0.18 * \text{oysters} + 0.12 * \text{POS-ADJ}\nonumber\\
& + 0.096 * \text{POS-NUM} + 0.096 * \text{POS-ADV}\nonumber\\
& + 0.089 * \text{POS-PROPN} + 0.086 * \text{tropical}\nonumber\\
& + 0.075 * \text{underwater} + 0.068 * \text{dunes}\nonumber\\
& + 0.063 * \text{seas} + 0.06 * \text{diver}\nonumber\\
& -0.058 * \text{PAST-TENSE
}\nonumber\\ & 
+ 0.042 * \text{sandstone}\nonumber\\
& -0.025 * \text{demon} + 0.02 * \text{marine}\nonumber\\
& -0.019 * \text{PARTICIPLE
}\nonumber\\ &  -0.014 * \text{POS-NOUN}\nonumber\\
& -0.012 * \text{french} -0.0041 * \text{witches}\nonumber
\end{align}
\begin{align}
&\text{Coulson} = 0.85 * \text{hacking}\nonumber\\
& + 0.72 * \text{C0} + 0.59 * \text{POS-PROPN}\nonumber\\
& + 0.25 * \text{CAPITALIZATION}\nonumber\\
& -0.23 * \text{POS-ADJ}\nonumber\\
& -0.19 * \text{POS-NOUN} + 0.17 * \text{butler}\nonumber\\
& -0.17 * \text{southern} + 0.15 * \text{POS-VERB}\nonumber\\
& -0.14 * \text{website} -0.12 * \text{com}\nonumber\\
& -0.12 * \text{roaring} + 0.1 * \text{solicitors}\nonumber\\
& -0.094 * \text{POS-ADV} + 0.074 * \text{oats}\nonumber\\
& -0.068 * \text{cathedral} -0.064 * \text{PARTICIPLE
}\nonumber\\
& + 0.061 * \text{inquiry} -0.06 * \text{dances}\nonumber\\
& -0.056 * \text{fan} + 0.042 * \text{POS-NUM}\nonumber\\
& -0.029 * \text{provinces} -0.029 * \text{finals}\nonumber\\
& -0.02 * \text{dance} -0.017 * \text{waters}\nonumber\\
& -0.013 * \text{tango} -0.013 * \text{shame}\nonumber\\
& -0.012 * \text{PAST-TENSE
}\nonumber\\
& -0.005 * \text{PLURAL-NOUN
}\nonumber
\end{align}
\begin{align}
&\text{roundabout} = 0.72 * \text{bypass}\nonumber\\
& + 0.4 * \text{roadway} -0.28 * \text{CAPITALIZATION}\nonumber\\
& + 0.22 * \text{plaza} -0.16 * \text{PLURAL-NOUN
}\nonumber\\
& + 0.11 * \text{POS-ADJ} + 0.11 * \text{clumsy}\nonumber\\
& + 0.1 * \text{airfield} + 0.088 * \text{POS-ADV}\nonumber\\
& -0.08 * \text{biological} + 0.079 * \text{C0}\nonumber\\
& + 0.051 * \text{POS-NOUN} + 0.043 * \text{PAST-TENSE
}\nonumber\\
& + 0.039 * \text{PARTICIPLE
} + 0.028 * \text{caravan}\nonumber\\
& + 0.025 * \text{ironic} + 0.021 * \text{POS-VERB}\nonumber\\
& -0.021 * \text{POS-NUM} -0.0038 * \text{POS-PROPN}\nonumber\\
& + 0.003 * \text{nonsensical}\nonumber
\end{align}
\begin{align}
&\text{Hub} = 0.49 * \text{bustling}\nonumber\\
& + 0.47 * \text{C0} + 0.4 * \text{portal}\nonumber\\
& + 0.39 * \text{infrastructure} + 0.32 * \text{POS-NOUN}\nonumber\\
& + 0.31 * \text{CAPITALIZATION} + 0.31 * \text{central}\nonumber\\
& -0.13 * \text{POS-PROPN} -0.1 * \text{PLURAL-NOUN
}\nonumber\\
& + 0.069 * \text{outage} + 0.068 * \text{centre}\nonumber\\
& + 0.061 * \text{POS-NUM} + 0.058 * \text{connectivity}\nonumber\\
& + 0.058 * \text{PARTICIPLE
} -0.057 * \text{POS-ADJ}\nonumber\\
& + 0.043 * \text{PAST-TENSE
} -0.043 * \text{POS-VERB}\nonumber\\
& + 0.027 * \text{POS-ADV}\nonumber
\end{align}
\begin{align}
&\text{environmental} = 0.43 * \text{sustainability}\nonumber\\
& + 0.43 * \text{economic} + 0.38 * \text{POS-ADJ}\nonumber\\
& -0.3 * \text{CAPITALIZATION} -0.27 * \text{POS-VERB}\nonumber\\
& + 0.27 * \text{regulatory} -0.2 * \text{PAST-TENSE
}\nonumber\\
& + 0.18 * \text{biological} + 0.17 * \text{campaigner}\nonumber\\
& + 0.17 * \text{POS-NUM} + 0.14 * \text{thermal}\nonumber\\
& -0.14 * \text{POS-ADV} -0.12 * \text{POS-NOUN}\nonumber\\
& + 0.1 * \text{health} -0.1 * \text{C0}\nonumber\\
& + 0.087 * \text{PARTICIPLE
} + 0.087 * \text{POS-PROPN}\nonumber\\
& -0.084 * \text{PLURAL-NOUN
} + 0.073 * \text{outdoor}\nonumber\\
& + 0.055 * \text{chemical} + 0.0055 * \text{cultural}\nonumber
\end{align}
\begin{align}
&\text{Churchill} = 0.84 * \text{wartime}\nonumber\\
& + 0.6 * \text{C0} + 0.41 * \text{CAPITALIZATION}\nonumber\\
& + 0.4 * \text{quotation} + 0.38 * \text{POS-PROPN}\nonumber\\
& + 0.36 * \text{statesman} -0.21 * \text{PARTICIPLE
}\nonumber\\
& -0.14 * \text{astronomer} -0.14 * \text{POS-NOUN}\nonumber\\
& -0.11 * \text{POS-ADJ} -0.1 * \text{PAST-TENSE
}\nonumber\\
& + 0.082 * \text{POS-VERB} + 0.078 * \text{POS-NUM}\nonumber\\
& + 0.064 * \text{POS-ADV} + 0.045 * \text{advising}\nonumber\\
& -0.025 * \text{architectures} + 0.022 * \text{PLURAL-NOUN
}\nonumber\\
& + 0.017 * \text{pint} + 0.013 * \text{fascism}\nonumber
\end{align}
\begin{align}
&\text{resident} = 0.54 * \text{citizens}\nonumber\\
& + 0.49 * \text{native} + 0.37 * \text{visiting}\nonumber\\
& -0.19 * \text{PLURAL-NOUN
} + 0.12 * \text{PAST-TENSE
}\nonumber\\
& + 0.11 * \text{caretaker} -0.099 * \text{CAPITALIZATION}\nonumber\\
& -0.094 * \text{C0} + 0.082 * \text{PARTICIPLE
}\nonumber\\
& + 0.082 * \text{ward} + 0.077 * \text{POS-NOUN}\nonumber\\
& -0.039 * \text{POS-ADV} + 0.022 * \text{proprietor}\nonumber\\
& + 0.022 * \text{POS-VERB} -0.0065 * \text{POS-NUM}\nonumber\\
& + 0.0045 * \text{POS-PROPN} + 0.0036 * \text{POS-ADJ}\nonumber
\end{align}
\begin{align}
&\text{backers} = 0.64 * \text{sponsors}\nonumber\\
& + 0.4 * \text{POS-NOUN} -0.4 * \text{CAPITALIZATION}\nonumber\\
& + 0.33 * \text{advocates} + 0.28 * \text{PLURAL-NOUN
}\nonumber\\
& + 0.19 * \text{POS-PROPN} + 0.18 * \text{businessman}\nonumber\\
& + 0.18 * \text{businessmen} + 0.16 * \text{fans}\nonumber\\
& -0.15 * \text{POS-ADJ} + 0.12 * \text{PARTICIPLE
}\nonumber\\
& + 0.12 * \text{PAST-TENSE
} -0.12 * \text{POS-ADV}\nonumber\\
& + 0.092 * \text{whose} + 0.082 * \text{opposition}\nonumber\\
& + 0.065 * \text{POS-VERB} + 0.056 * \text{candidacy}\nonumber\\
& + 0.055 * \text{touted} + 0.047 * \text{startups}\nonumber\\
& -0.024 * \text{POS-NUM} + 0.024 * \text{rebels}\nonumber\\
& + 0.014 * \text{reformist} + 0.013 * \text{investment}\nonumber\\
& -0.002 * \text{C0}\nonumber
\end{align}
\begin{align}
&\text{rudimentary} = 0.84 * \text{basics}\nonumber\\
& -0.65 * \text{C0} + 0.49 * \text{POS-ADJ}\nonumber\\
& -0.41 * \text{POS-VERB} + 0.41 * \text{apparatus}\nonumber\\
& -0.36 * \text{POS-NOUN} + 0.35 * \text{improvised}\nonumber\\
& + 0.15 * \text{POS-ADV} + 0.099 * \text{CAPITALIZATION}\nonumber\\
& + 0.072 * \text{PARTICIPLE
} + 0.069 * \text{PLURAL-NOUN
}\nonumber\\
& + 0.062 * \text{POS-NUM} + 0.059 * \text{develop}\nonumber\\
& + 0.05 * \text{PAST-TENSE
} + 0.043 * \text{POS-PROPN}\nonumber
\end{align}
\begin{align}
&\text{admire} = 0.73 * \text{admirable}\nonumber\\
& -0.66 * \text{PARTICIPLE
} -0.65 * \text{C0}\nonumber\\
& + 0.31 * \text{magnificent} + 0.23 * \text{CAPITALIZATION}\nonumber\\
& + 0.16 * \text{criticize} + 0.16 * \text{POS-NOUN}\nonumber\\
& -0.16 * \text{PAST-TENSE
} + 0.14 * \text{loves}\nonumber\\
& + 0.1 * \text{POS-PROPN} + 0.1 * \text{beauty}\nonumber\\
& -0.098 * \text{POS-NUM} -0.068 * \text{PLURAL-NOUN
}\nonumber\\
& + 0.066 * \text{devotion} -0.061 * \text{POS-ADV}\nonumber\\
& -0.058 * \text{POS-ADJ} + 0.039 * \text{openness}\nonumber\\
& + 0.02 * \text{charming} -0.00015 * \text{POS-VERB}\nonumber
\end{align}
\begin{align}
&\text{re-add} = -0.65 * \text{PARTICIPLE
}\nonumber\\
& -0.47 * \text{POS-NUM} + 0.44 * \text{deleted}\nonumber\\
& + 0.43 * \text{POS-VERB} + 0.41 * \text{cruft}\nonumber\\
& -0.41 * \text{C0} -0.3 * \text{PAST-TENSE
}\nonumber\\
& + 0.28 * \text{section} + 0.19 * \text{CAPITALIZATION}\nonumber\\
& + 0.17 * \text{categorization} + 0.16 * \text{unblock}\nonumber\\
& + 0.15 * \text{POS-ADV} + 0.11 * \text{POS-PROPN}\nonumber\\
& + 0.098 * \text{reversion} -0.09 * \text{POS-ADJ}\nonumber\\
& + 0.09 * \text{POS-NOUN} + 0.061 * \text{inserting}\nonumber\\
& + 0.046 * \text{reference} + 0.043 * \text{sourcing}\nonumber\\
& + 0.034 * \text{template} + 0.027 * \text{encyclopedic}\nonumber\\
& + 0.013 * \text{modify} -0.0088 * \text{battleship}\nonumber\\
& -0.0071 * \text{cow} + 0.006 * \text{PLURAL-NOUN
}\nonumber
\end{align}
\begin{align}
&\text{visuals} = 0.47 * \text{cinematography}\nonumber\\
& -0.47 * \text{CAPITALIZATION} + 0.29 * \text{evocative}\nonumber\\
& + 0.25 * \text{multimedia} + 0.21 * \text{videos}\nonumber\\
& + 0.19 * \text{POS-NOUN} + 0.15 * \text{PLURAL-NOUN
}\nonumber\\
& + 0.14 * \text{POS-PROPN} + 0.12 * \text{hallucinations}\nonumber\\
& + 0.11 * \text{awesome} + 0.1 * \text{PARTICIPLE
}\nonumber\\
& + 0.08 * \text{video} -0.079 * \text{POS-VERB}\nonumber\\
& + 0.076 * \text{sounds} + 0.076 * \text{POS-ADJ}\nonumber\\
& + 0.076 * \text{slick} + 0.075 * \text{POS-ADV}\nonumber\\
& + 0.066 * \text{C0} + 0.062 * \text{dazzling}\nonumber\\
& + 0.052 * \text{colorful} + 0.044 * \text{interactive}\nonumber\\
& + 0.027 * \text{jarring} + 0.019 * \text{visualization}\nonumber\\
& + 0.0047 * \text{PAST-TENSE
} + 0.00025 * \text{POS-NUM}\nonumber
\end{align}
\begin{align}
&\text{Conflict} = 0.61 * \text{POS-NOUN}\nonumber\\
& -0.49 * \text{POS-PROPN} + 0.44 * \text{warfare}\nonumber\\
& + 0.4 * \text{escalation} + 0.4 * \text{peace}\nonumber\\
& + 0.36 * \text{C0} + 0.24 * \text{guideline}\nonumber\\
& + 0.23 * \text{CAPITALIZATION} + 0.21 * \text{PARTICIPLE
}\nonumber\\
& + 0.19 * \text{ethnic} -0.19 * \text{POS-VERB}\nonumber\\
& + 0.16 * \text{resolved} + 0.12 * \text{POS-NUM}\nonumber\\
& -0.099 * \text{PLURAL-NOUN
} + 0.078 * \text{PAST-TENSE
}\nonumber\\
& + 0.07 * \text{divergence} + 0.065 * \text{geopolitical}\nonumber\\
& -0.05 * \text{stationary} + 0.038 * \text{POS-ADJ}\nonumber\\
& -0.032 * \text{shops} + 0.03 * \text{polarized}\nonumber\\
& -0.012 * \text{POS-ADV}\nonumber
\end{align}
\begin{align}
&\text{hitter} = -0.54 * \text{CAPITALIZATION}\nonumber\\
& + 0.45 * \text{C0} -0.42 * \text{PLURAL-NOUN
}\nonumber\\
& + 0.42 * \text{shortstop} + 0.36 * \text{designated}\nonumber\\
& + 0.32 * \text{batting} + 0.3 * \text{POS-VERB}\nonumber\\
& + 0.21 * \text{POS-NOUN} + 0.18 * \text{POS-ADV}\nonumber\\
& + 0.17 * \text{pitchers} + 0.17 * \text{pitcher}\nonumber\\
& + 0.14 * \text{catcher} -0.12 * \text{PARTICIPLE
}\nonumber\\
& -0.1 * \text{POS-NUM} -0.096 * \text{inane}\nonumber\\
& + 0.087 * \text{guy} + 0.073 * \text{POS-PROPN}\nonumber\\
& + 0.048 * \text{exert} -0.014 * \text{PAST-TENSE
}\nonumber\\
& + 0.0071 * \text{outs} -0.0064 * \text{POS-ADJ}\nonumber\\
& + 0.0019 * \text{swings}\nonumber
\end{align}
\begin{align}
&\text{fence} = 0.52 * \text{wire}\nonumber\\
& + 0.43 * \text{gates} -0.41 * \text{CAPITALIZATION}\nonumber\\
& + 0.35 * \text{yard} -0.32 * \text{PLURAL-NOUN
}\nonumber\\
& + 0.21 * \text{shrubs} + 0.14 * \text{barn}\nonumber\\
& + 0.14 * \text{ditch} + 0.09 * \text{POS-VERB}\nonumber\\
& + 0.085 * \text{side} -0.07 * \text{PARTICIPLE
}\nonumber\\
& -0.052 * \text{POS-ADJ} + 0.042 * \text{POS-NUM}\nonumber\\
& -0.032 * \text{POS-PROPN} -0.02 * \text{PAST-TENSE
}\nonumber\\
& -0.012 * \text{C0} + 0.0068 * \text{nailed}\nonumber\\
& -0.0047 * \text{POS-ADV} + 0.00013 * \text{POS-NOUN}\nonumber
\end{align}
\begin{align}
&\text{1978} = 0.97 * \text{1970s}\nonumber\\
& -0.89 * \text{POS-ADJ} -0.6 * \text{POS-PROPN}\nonumber\\
& + 0.49 * \text{POS-NUM} -0.42 * \text{POS-NOUN}\nonumber\\
& + 0.21 * \text{C0} -0.18 * \text{PLURAL-NOUN
}\nonumber\\
& -0.12 * \text{POS-ADV} -0.081 * \text{PARTICIPLE
}\nonumber\\
& -0.073 * \text{CAPITALIZATION}\nonumber\\ &  + 0.067 * \text{POS-VERB}\nonumber\\
& + 0.041 * \text{PAST-TENSE
} + 0.039 * \text{seventies}\nonumber\\
& + 0.026 * \text{contends}\nonumber
\end{align}
\begin{align}
&\text{heroine} = 0.66 * \text{hero}\nonumber\\
& + 0.35 * \text{protagonist}\nonumber\\ &  -0.34 * \text{CAPITALIZATION}\nonumber\\
& -0.25 * \text{PLURAL-NOUN
} + 0.14 * \text{actress}\nonumber\\
& + 0.13 * \text{girl} + 0.1 * \text{C0}\nonumber\\
& + 0.1 * \text{PAST-TENSE
} + 0.071 * \text{POS-PROPN}\nonumber\\
& + 0.07 * \text{POS-NOUN} + 0.063 * \text{POS-ADV}\nonumber\\
& -0.058 * \text{POS-NUM} + 0.051 * \text{protagonists}\nonumber\\
& -0.029 * \text{POS-VERB} + 0.026 * \text{PARTICIPLE
}\nonumber\\
& + 0.015 * \text{POS-ADJ} + 0.014 * \text{goddess}\nonumber
\end{align}
\begin{align}
&\text{structure} = 0.91 * \text{structures}\nonumber\\
& -0.35 * \text{CAPITALIZATION}\nonumber\\ &  -0.25 * \text{PLURAL-NOUN
}\nonumber\\
& + 0.17 * \text{structuring} + 0.16 * \text{POS-NOUN}\nonumber\\
& -0.085 * \text{POS-VERB} -0.078 * \text{PAST-TENSE
}\nonumber\\
& + 0.05 * \text{POS-ADV} -0.039 * \text{POS-ADJ}\nonumber\\
& -0.034 * \text{POS-PROPN} + 0.029 * \text{POS-NUM}\nonumber\\
& + 0.026 * \text{structural} + 0.022 * \text{reorganization}\nonumber\\
& -0.022 * \text{C0} + 0.0079 * \text{PARTICIPLE
}\nonumber
\end{align}
\begin{align}
&\text{wizards} = 0.65 * \text{magic}\nonumber\\
& + 0.41 * \text{witches}\nonumber\\ &  -0.41 * \text{CAPITALIZATION}\nonumber\\
& + 0.36 * \text{PLURAL-NOUN
}\nonumber\\ &  + 0.19 * \text{POS-NOUN}\nonumber\\
& -0.19 * \text{POS-NUM} + 0.15 * \text{POS-ADJ}\nonumber\\
& + 0.15 * \text{tech} + 0.13 * \text{POS-ADV}\nonumber\\
& + 0.098 * \text{PAST-TENSE
} + 0.09 * \text{wannabe}\nonumber\\
& + 0.084 * \text{dragons} + 0.084 * \text{knights}\nonumber\\
& + 0.052 * \text{C0} -0.036 * \text{POS-PROPN}\nonumber\\
& + 0.022 * \text{PARTICIPLE
} + 0.015 * \text{err}\nonumber\\
& -0.013 * \text{POS-VERB} + 0.0041 * \text{guru}\nonumber
\end{align}
\begin{align}
&\text{autistic} = 0.49 * \text{preschool}\nonumber\\
& + 0.37 * \text{epilepsy} -0.35 * \text{POS-NOUN}\nonumber\\
& + 0.33 * \text{POS-ADJ} -0.25 * \text{papal}\nonumber\\
& + 0.22 * \text{son} + 0.21 * \text{POS-PROPN}\nonumber\\
& + 0.16 * \text{twins} + 0.12 * \text{PAST-TENSE
}\nonumber\\
& + 0.12 * \text{therapist} + 0.12 * \text{PARTICIPLE
}\nonumber\\
& -0.1 * \text{CAPITALIZATION} + 0.084 * \text{teenage}\nonumber\\
& + 0.072 * \text{trait} + 0.069 * \text{psychologist}\nonumber\\
& + 0.067 * \text{behaviors} + 0.056 * \text{kid}\nonumber\\
& + 0.054 * \text{hospitalized} -0.053 * \text{C0}\nonumber\\
& + 0.052 * \text{manipulative} -0.038 * \text{POS-NUM}\nonumber\\
& + 0.037 * \text{granddaughter} -0.03 * \text{rounds}\nonumber\\
& + 0.03 * \text{PLURAL-NOUN
} + 0.027 * \text{campers}\nonumber\\
& + 0.026 * \text{POS-VERB} + 0.0092 * \text{dementia}\nonumber\\
& -0.0054 * \text{regional} -0.0052 * \text{POS-ADV}\nonumber\\
& -0.0023 * \text{sedan}\nonumber
\end{align}
\begin{align}
&\text{tornado} = 0.72 * \text{hurricane}\nonumber\\
& + 0.49 * \text{C0} -0.47 * \text{CAPITALIZATION}\nonumber\\
& + 0.28 * \text{typhoon} -0.26 * \text{PLURAL-NOUN
}\nonumber\\
& + 0.23 * \text{tractor} + 0.21 * \text{POS-VERB}\nonumber\\
& + 0.16 * \text{POS-ADJ} + 0.12 * \text{POS-NUM}\nonumber\\
& + 0.1 * \text{flattened} -0.1 * \text{ports}\nonumber\\
& -0.097 * \text{opium} + 0.072 * \text{POS-ADV}\nonumber\\
& + 0.053 * \text{POS-PROPN} + 0.053 * \text{avalanche}\nonumber\\
& + 0.052 * \text{tape} + 0.05 * \text{earthquake}\nonumber\\
& -0.045 * \text{colonial} -0.043 * \text{POS-NOUN}\nonumber\\
& + 0.028 * \text{musical} -0.025 * \text{handsets}\nonumber\\
& + 0.014 * \text{terrifying} + 0.013 * \text{PAST-TENSE
}\nonumber\\
& + 0.013 * \text{occurrences} -0.0067 * \text{labour}\nonumber\\
& + 0.0025 * \text{PARTICIPLE
}\nonumber
\end{align}
\begin{align}
&\text{1852} = 0.85 * \text{1800s}\nonumber\\
& -0.81 * \text{POS-PROPN} -0.78 * \text{POS-ADJ}\nonumber\\
& -0.61 * \text{POS-NOUN} + 0.45 * \text{POS-NUM}\nonumber\\
& + 0.43 * \text{C0} -0.31 * \text{CAPITALIZATION}\nonumber\\
& + 0.29 * \text{renders} + 0.25 * \text{POS-VERB}\nonumber\\
& -0.19 * \text{PLURAL-NOUN
} + 0.16 * \text{noted}\nonumber\\
& -0.15 * \text{PARTICIPLE
} + 0.13 * \text{underscored}\nonumber\\
& -0.082 * \text{POS-ADV} + 0.063 * \text{insisting}\nonumber\\
& + 0.029 * \text{PAST-TENSE
}\nonumber
\end{align}
\begin{align}
&\text{gloom} = 0.66 * \text{gloomy}\nonumber\\
& -0.46 * \text{CAPITALIZATION} -0.32 * \text{POS-VERB}\nonumber\\
& + 0.28 * \text{pessimism} + 0.28 * \text{darkness}\nonumber\\
& + 0.15 * \text{PAST-TENSE
} -0.14 * \text{PLURAL-NOUN
}\nonumber\\
& + 0.14 * \text{POS-NOUN} + 0.12 * \text{PARTICIPLE
}\nonumber\\
& + 0.11 * \text{POS-NUM} -0.058 * \text{C0}\nonumber\\
& -0.058 * \text{POS-ADV} + 0.052 * \text{misery}\nonumber\\
& + 0.051 * \text{POS-PROPN} + 0.037 * \text{slump}\nonumber\\
& -0.025 * \text{POS-ADJ}\nonumber
\end{align}
\begin{align}
&\text{recycle} = -0.65 * \text{PARTICIPLE
}\nonumber\\
& + 0.61 * \text{bin} + 0.49 * \text{rubbish}\nonumber\\
& -0.48 * \text{C0} -0.47 * \text{PAST-TENSE
}\nonumber\\
& + 0.27 * \text{POS-VERB} + 0.18 * \text{POS-NOUN}\nonumber\\
& + 0.15 * \text{utilize} + 0.14 * \text{plastic}\nonumber\\
& -0.12 * \text{POS-NUM} + 0.1 * \text{excess}\nonumber\\
& + 0.088 * \text{sustainability} + 0.083 * \text{POS-PROPN}\nonumber\\
& + 0.066 * \text{aluminum} + 0.042 * \text{gramthesize}\nonumber\\
& -0.037 * \text{PLURAL-NOUN
} + 0.036 * \text{POS-ADJ}\nonumber\\
& + 0.035 * \text{refurbished} + 0.034 * \text{POS-ADV}\nonumber\\
& + 0.03 * \text{converter} + 0.026 * \text{nitrogen}\nonumber\\
& -0.004 * \text{CAPITALIZATION} + 0.0024 * \text{saving}\nonumber
\end{align}
\end{document}